\documentclass{article}

\usepackage{microtype}
\usepackage{graphicx}
\usepackage{subcaption}
\usepackage{booktabs} % for professional tables

\usepackage[accepted]{icml2026}

\usepackage{amssymb}
\usepackage{mathtools}
\usepackage{amsthm}
\usepackage{hyperref}

\usepackage[utf8]{inputenc}
\usepackage[T1]{fontenc}

\usepackage{amsfonts}
\usepackage{amsmath}
\usepackage{nicefrac}
\usepackage{xcolor}
\usepackage{siunitx}
\usepackage{enumitem}
\usepackage{multirow,tabularx,threeparttable}
\usepackage{float}           % for [H] placement if you really need it

\DeclareUnicodeCharacter{202F}{\,}

\usepackage[capitalize,noabbrev]{cleveref}

\theoremstyle{plain}

\theoremstyle{definition}

\theoremstyle{remark}

\icmltitlerunning{AutoMat: Physics-Guided Agentic Reasoning for Inverse Microscopy}
\icmlsetsymbol{equal}{*}
\hypersetup{
  pdftitle={AutoMat: Physics-Guided Agentic Reasoning for Solving Ill-Posed Inverse Microscopy Problems},
  pdfauthor={Yaotian Yang, Yiwen Tang, Yizhe Chen, Xiao Chen, Jiangjie Qiu, Hao Xiong, Haoyu Yin, Zhiyao Luo, Yifei Zhang, Sijia Tao, Wentao Li, Qinghua Zhang, Yuqiang Li, Wanli Ouyang, Bin Zhao, Xiaonan Wang, Fei Wei}
}
\begin{document}

\twocolumn[
\icmltitle{AutoMat: Physics-Guided Agentic Reasoning for Solving Ill-Posed Inverse Microscopy Problems}

\icmlkeywords{AI for Science, Electron Microscopy, Agentic Tool Use, Crystal Structure Reconstruction}

\begin{icmlauthorlist}
  \icmlauthor{Yaotian Yang}{equal,tsinghua}
  \icmlauthor{Yiwen Tang}{equal,shailab}
  \icmlauthor{Yizhe Chen}{equal,tsinghua}
  \icmlauthor{Xiao Chen}{tsinghua,ordoslab}
  \icmlauthor{Jiangjie Qiu}{tsinghua}
  \icmlauthor{Hao Xiong}{tsinghua}
  \icmlauthor{Haoyu Yin}{tsinghua}
  \icmlauthor{Zhiyao Luo}{tsinghua}
  \icmlauthor{Yifei Zhang}{tsinghua}
  \icmlauthor{Sijia Tao}{tsinghua}
  \icmlauthor{Wentao Li}{tsinghua}
  \icmlauthor{Qinghua Zhang}{tsinghua}
  \icmlauthor{Yuqiang Li}{shailab}
  \icmlauthor{Wanli Ouyang}{shailab}
  \icmlauthor{Bin Zhao}{shailab}
  \icmlauthor{Xiaonan Wang}{tsinghua}
  \icmlauthor{Fei Wei}{tsinghua,ordoslab}
\end{icmlauthorlist}

\icmlaffiliation{tsinghua}{Department of Chemical Engineering, Tsinghua University, Beijing, China}
\icmlaffiliation{shailab}{Shanghai Artificial Intelligence Laboratory, Shanghai, China}
\icmlaffiliation{ordoslab}{Ordos Laboratory, Ordos, China}

\icmlcorrespondingauthor{Xiao Chen}{chenx123@tsinghua.edu.cn}
\icmlcorrespondingauthor{Bin Zhao}{bin@nwpu.edu.cn}
\icmlcorrespondingauthor{Xiaonan Wang}{wangxiaonan@tsinghua.edu.cn}
\icmlcorrespondingauthor{Fei Wei}{wf-dce@tsinghua.edu.cn}

\vskip 0.3in
]

% Required even if empty:
\printAffiliationsAndNotice{\icmlEqualContribution}

\begin{abstract}
Reconstructing atomistic crystal structures from a single noisy STEM projection is an ill-posed inverse problem: multiple lattices can explain similar contrast, and purely feed-forward models cannot verify physical validity. We present \textbf{AutoMat}, a failure-aware agentic \emph{controller} that performs inference-time hypothesis search with \emph{closed-loop verification} to convert Scanning Transmission Electron Microscopy (STEM) images into simulation-ready crystal structures and downstream properties. AutoMat composes perception and physics modules---pattern-adaptive denoising, physics-guided template retrieval \emph{as a state-dependent auxiliary branch}, symmetry-constrained atomic reconstruction, and MLIP-based relaxation/validation---and triggers rollback-and-retry when verification fails. For systematic evaluation, we introduce \textbf{STEM2Mat-Bench}, a benchmark dataset containing 450+ annotated samples. Performance is assessed using lattice root-mean-square deviation (RMSD), formation energy mean absolute error (MAE), and structure matching accuracy. Results demonstrate that AutoMat outperforms existing approaches including SOTA models, specialized domain tools, and closed-source multimodal large models. This work establishes a direct pathway from microscopic characterization to atomic-scale modeling, addressing a fundamental challenge in materials science.
\end{abstract}

\section{Introduction}
\label{intro}

\begin{figure}[t]
  \centering
  \includegraphics[width=0.94\linewidth]{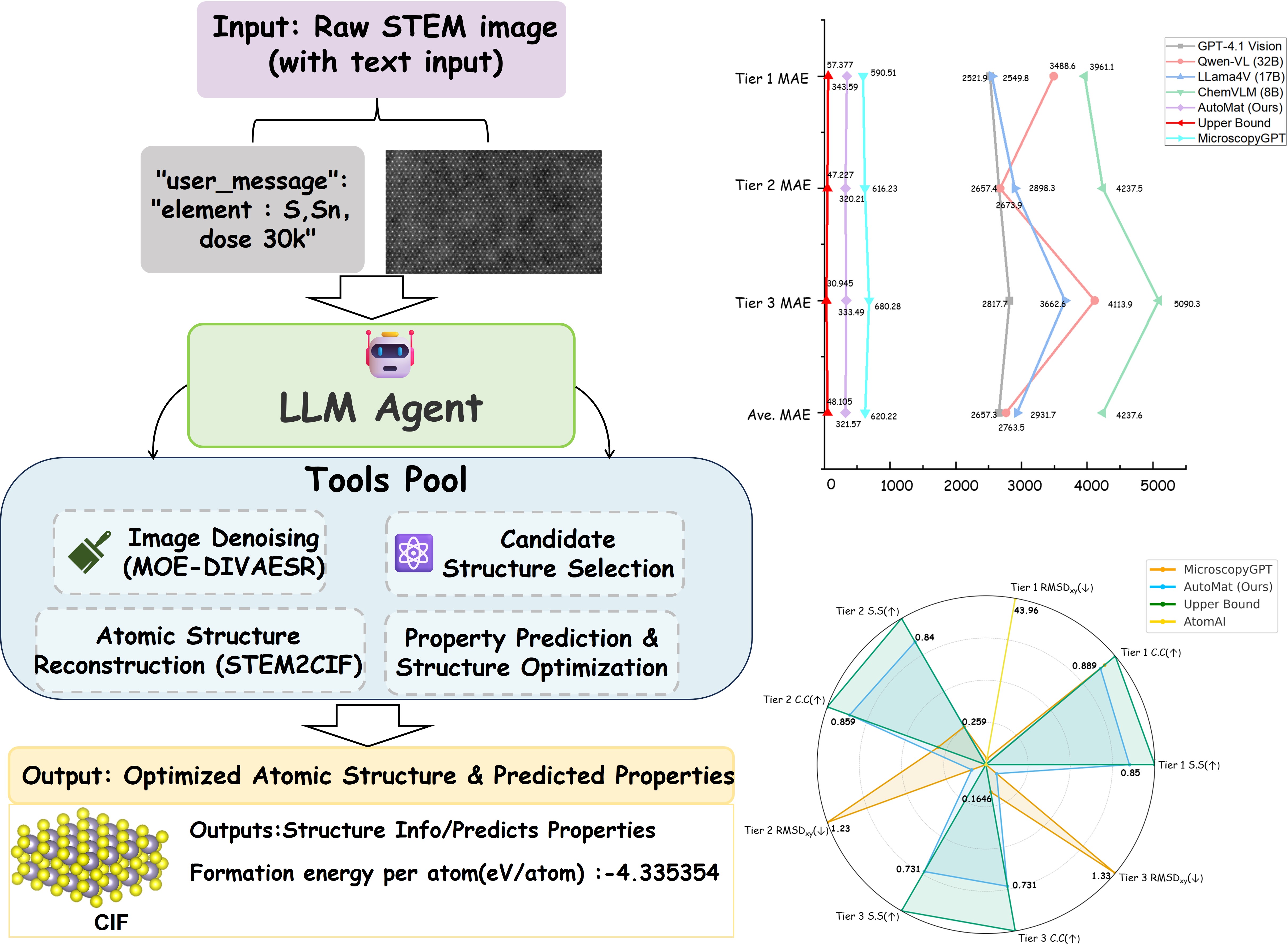}
  \caption{\textbf{Overview of AutoMat.} The left part shows a failure-aware reasoning \emph{controller} that composes four modules—pattern-adaptive denoising, template selection (state-triggered auxiliary branch), atomic reconstruction (STEM2CIF), and MLIP-based relaxation/property prediction—into a closed-loop verification workflow. The right panel presents a line chart and radar plot comparing different models in terms of energy and structural errors.}
  \label{fig:teaser}
\vspace{-0.4cm}
\end{figure}

Machine learning interatomic potentials and force fields now approach ~\emph{ab initio} accuracy in predicting atomic energies and forces~\citep{yang2024mattersimdeeplearningatomistic,batatia2024foundationmodelatomisticmaterials,gasteiger2021gemnet,liao2024equiformerv2improvedequivarianttransformer,lu2023highlyaccuratequantumchemical} but remain limited by experimental validated crystal structures.
Meanwhile, Scanning Transmission Electron Microscopy (STEM) can image atoms at sub-ångström resolution~\citep{RN46,RN47,RN80,RN87}, yet translating into quantitative crystal structures still relies on expert-driven, time-consuming annotation.
This creates a gap between structural representation and theoretical validation in materials science~\citep{RN28,kalinin2023machine}.

Although recent advances in STEM image analysis have shown promise, most existing studies focus on individual components like denoising~\citep{https://doi.org/10.1002/advs.202408629,RN32,RN93,RN92}, atom localization~\citep{RN36,borshon2024predicting,eliasson2024localization}, reconstruction~\citep{de2022experimental,stoops2025obtaining,huang2025auto} and phase classification~\citep{RN30,RN64}. These approaches remain fragmented and are not integrated into end-to-end system. 
Conventional image denoising techniques suppress noise and improve contrast at the pixel level but cannot yield periodic or chemically meaningful crystal structures~\citep{RN5,RN94,RN92,https://doi.org/10.1002/advs.202408629}. 
Atomic detection models can localize atomic peaks but cannot infer complete lattices\footnote{In crystallography, a lattice refers to the periodic arrangement of atoms in a crystal structure.} or identify atomic species.
General-purpose multimodal models like GPT-4.1mini~\citep{achiam2023gpt} and Qwen2.5-VL~\citep{bai2025qwen2} exhibit basic image understanding but lack the ability to produce simulation-ready formats such as crystallographic file (CIF). Even domain-specific tools like AtomAI~\citep{ziatdinov2022atomai} and the newly released domain SOTA model MicroscopyGPT~\citep{doi:10.1021/acs.jpclett.5c01257} can only predict atomic coordinates and structural descriptions for minimal systems, without supporting complete CIF structure reconstruction or property prediction. Meanwhile, public datasets (e.g., JARVIS-STM) mainly target STM (Scanning Tunneling Microscopy) images, lack DFT-level energy labels, and are unsuitable for benchmarking structure–property pipelines.
As a result, the field lacks a fully automated end-to-end system that can convert raw STEM images into reconstructed structures and simulated properties, along with the standardized benchmark for comprehensive evaluation.

To address this gap, as shown in Fig.~\ref{fig:teaser}, we introduce \textbf{AutoMat}, a failure-aware agentic \emph{controller} that links STEM imaging with atomistic simulation via inference-time hypothesis search and closed-loop verification, along with a benchmark tailored to the task.
At the core of AutoMat is a controller equipped with a suite of four modular tools, which it can dynamically compose to solve the task:
1) \textbf{Pattern-Adaptive STEM Image denoising}: We apply a pattern-adaptive mixture-of-experts network, MOE-Denoising Inference Variational Autoencoder Super-Resolution~\citep{https://doi.org/10.1002/advs.202408629}(MOE-DIVAESR), to denoise and enhance raw iDPC-STEM\footnote{iDPC-STEM stands for integrated Differential Phase Contrast Scanning Transmission Electron Microscopy, a technique that enhances light element contrast in atomic-resolution imaging.} images. 
The ResNet-18-based gating network elects the most suitable expert network for each input image based on its estimated noise level, enabling joint denoising, inpainting, and super-resolution.
2) \textbf{Physics-Guided Template Retrieval}: Enhanced images are matched to a large-scale library of simulated STEM projections. Top candidate structures are retrieved using pixel similarity and filtered by elemental contrast patterns
 to produce strong structural priors.
3) \textbf{Symmetry-Constrained Structure Reconstruction}: Atomic peaks are detected via unsupervised clustering. We fit the lattice under symmetry constraints, assign atomic species based on the candidate, and generate the standard CIF file representing the periodic crystal structure. 
4) \textbf{Energy Evaluation via Machine-Learned Potential}: The reconstructed structure is relaxed using the pretrained MatterSim potential to predict formation energy. 
The language agent autonomously orchestrates the workflow, planning tool execution and adaptively retrying failed steps based on quality checks.

To support rigorous evaluation, we curated 2,143 high-quality monolayer structures from C2DB~\citep{haastrup2018computational}, Materials Project~\citep{jain2013commentary}, and OpenCrystal~\citep{Vaitkus2023}, and simulated their corresponding iDPC-STEM images with abTEM~\citep{madsen2020abtem}. From this pool we selected 459 representative image–structure pairs, which constitute our \textbf{STEM2Mat} benchmark and are used for all subsequent evaluations. Our evaluation metrics include projected lattice RMSD, formation energy MAE, and atom-wise structure matching success rate. Additionally, we introduce three fine-grained indicators to assess reconstruction quality (e.g., atomic recall/precision), robustness across noise levels, and computational efficiency. 
On this benchmark, AutoMat achieves a projected RMSD of $0.11 \pm 0.03$~\AA{}, energy MAE below 350~meV/atom, and an atomic correspondence total success rate of 83.2\%, outperforming GPT-4.1-mini, Qwen-VL, LLama4V, ChemVLM~\citep{li2025chemvlm}, MicroscopyGPT and AtomAI by an order of magnitude.
These results demonstrate AutoMat as a reproducible and accurate end-to-end solution for microscopy-driven materials modeling.

The contributions of this paper are summarized as follows:

  \textbullet\ \textbf{Physics-Aware Reasoning Controller}.
  We formulate STEM-to-structure reconstruction as inference-time hypothesis search with \emph{closed-loop verification}, and introduce AutoMat as a controller that composes denoising, a state-dependent retrieval-guided prior branch, reconstruction, and relaxation with failure signals and rollback-and-retry.
  
  \textbullet\ \textbf{Algorithmic advancements}. We design MOE-DIVAESR as a pattern-adaptive denoiser for diverse STEM images, enabling efficient noise reduction, defect correction, and detail enhancement. We also develop STEM2CIF to reconstruct crystal structures by identifying the minimal repeating unit using symmetry heuristics and physical constraints, then converting the result into standard CIF format.
  
  \textbullet\ \textbf{STEM2Mat benchmark \& evaluation suite}. We curate 2,143 distinct monolayer structures to simulate large-field iDPC-STEM images, and provide a benchmark split with unambiguous evaluation protocols and unified metrics for lattice reconstruction accuracy, energy fidelity, robustness, and computational efficiency.

  \textbullet\ \textbf{Reproducible end-to-end results}. We show that AutoMat substantially improves both structural fidelity and downstream energy prediction compared with off-the-shelf multimodal baselines. Code, data, and evaluation tools are publicly available at \url{https://github.com/yyt-2378/AutoMat} and \url{https://huggingface.co/datasets/yaotianvector/STEM2Mat} to support reproducibility.

% =========================== 2 Related Work ===========================
\section{Related Work}
\label{sec:related_work}

\textbf{Automated Microscopy Image Analysis.}
In recent years, deep learning methods have been extensively applied to electron microscopy and scanning probe microscopy data analysis~\citep{https://doi.org/10.1002/advs.202408629,RN32,RN93,RN92,RN36,borshon2024predicting,de2022experimental,stoops2025obtaining,huang2025auto}.
Current approaches range from unsupervised defect detection to supervised atomic column identification.
AtomAI~\citep{ziatdinov2022atomai}, for example, integrates microscopy images with computational simulations, but primarily focuses on atom segmentation and identification.
More recently, SciLink~\citep{yao2025operationalizingserendipitymultiagentai} proposes an open-source \emph{multi-agent} framework that closes the loop between microscopy/spectroscopy experiments, literature-based novelty assessment, and theory-in-the-loop simulations, and has demonstrated ~~impressive~~ performance in automated defect localization and atomic-scale analysis.
However, its primary focus is on serendipity-aware, ~~high-level~~ experiment--theory orchestration.
It does not yet provide a ~~dedicated~~ solution for reconstructing electron-microscopy images into explicit crystal structures that can be quantitatively compared with theoretical models, nor for using such reconstructed structures as ~~direct~~ inputs to downstream simulations and property prediction.

\textbf{STEM Image to Structure Reconstruction.}
Existing methods for reconstructing crystal structures from STEM images typically rely on multiple images, prior structural information, or are limited to single-element systems.
For instance, De Backer et al.~\citep{de2022experimental} employed Bayesian algorithms for 3D reconstruction, but their approach was demonstrated only for simple single-element systems.
Currently, few methods can generate standard crystallographic information files (CIF) from single experimental images, especially for complex multi-element 2D crystals~\citep{stoops2025obtaining}.

\textbf{Vision-Language Models in Chemistry.}
Recently, multimodal large language models (ChemVLM~\citep{li2025chemvlm}, GPT-4.1mini~\citep{achiam2023gpt}, and Qwen-VL~\citep{bai2025qwen2}) have started to be explored in chemistry and materials science.
However, these models generally lack the capability to accurately handle detailed spatial structure tasks, and recent benchmarks further reveal that vision-language models can exhibit erroneous reasoning in visually grounded tasks~\citep{shi2026mmerror}.
Existing chemical agent tools (e.g., ChemCrow~\citep{bran2023chemcrowaugmentinglargelanguagemodels} and Chemagents~\citep{tang2025chemagentselfupdatinglibrarylarge}) are limited to text-based descriptions and cannot process image-based inputs, significantly restricting their applicability in microscopy-based analyses.
The field’s SOTA multimodal model MicroscopyGPT~\citep{doi:10.1021/acs.jpclett.5c01257} can generate structural descriptions from STEM images but cannot yet reconstruct CIFs or predict properties.

\textbf{Material Property Prediction Models.}
Advances in machine learning-based interatomic potentials (such as MatterSim~\citep{yang2024mattersimdeeplearningatomistic}, M3GNet~\citep{chen2022universal}, and MACE~\citep{batatia2023macehigherorderequivariant}) have significantly improved the accuracy of computational property predictions.
Combining these models with experimental imaging provides a novel, digital twin-like approach for validating structural reconstructions.
However, existing benchmarks predominantly evaluate models using theoretical datasets, lacking end-to-end assessments starting from experimental image inputs.

In summary, these approaches highlight progress and limitations in microscopy image analysis, structure reconstruction, multimodal modeling, and property prediction.
\textbf{AutoMat addresses these gaps} by integrating an agent from STEM images to material property prediction and establishes the STEM2Mat benchmark for evaluating the robustness and scalability of automated material characterization.

% =========================== 3 Dataset Construction ===========================
% =========================== 3 Dataset Construction ===========================
\section{Dataset Construction}
\label{sec:dataset}

\subsection{Composition and Taxonomy}
Our benchmark focuses on two-dimensional (2D) materials, whose atomic-scale
thickness allows STEM to resolve individual columns with minimal
multiple-scattering artifacts.
Starting from nearly 10,000 candidate structures harvested from C2DB,
Materials Project, and OpenCrystal,\footnote{All sources were queried in
April~2025 using identical monolayer filters.}
we followed a two-stage curation process.
First, automated filters removed non-stoichiometric, partially occupied, or 3D
bulk entries.
Second, domain experts inspected symmetry, cleavage energy, and substrate
feasibility, yielding \textbf{2,143} high-confidence monolayer crystals.
The collection spans six chemical families
(Fig.~\ref{fig:data_overview_appendix}):
\emph{(i)} classic 2D materials---graphene, MoS\textsubscript{2}, h-BN, black
phosphorus;
\emph{(ii)} emergent allotropes, e.g., silicene, borophene;
\emph{(iii)} conductive MXenes (23 distinct formulas);
\emph{(iv)} intrinsic 2D magnets such as CrI\textsubscript{3} and
Fe\textsubscript{3}GeTe\textsubscript{2};
\emph{(v)} Janus structures typified by MoSSe;
and \emph{(vi)} Ruddlesden--Popper--type 2D perovskites.
Elemental diversity is broad: 67 unique elements appear, producing 76 unary,
1,409 binary, and 658 ternary systems.
Each structure is stored as a \textsc{cif} file with validated lattice vectors
and fractional atomic coordinates.

\subsection{Image Simulation and Data Augmentation}
To simulate realistic large-field STEM imaging conditions, we generated
synthetic iDPC-STEM micrographs using the open-source \texttt{abTEM} simulation
engine.
For each structure, a random $12\times12$ to $16\times16$ supercell was
constructed and projected at \SI{0.1}{\angstrom/pixel} resolution.
Five electron-dose settings
($1\text{--}6\times10^{4}\,\mathrm{e}^{-}/\si{\angstrom^{2}}$) and realistic
lens aberrations were sampled to mimic experimental conditions.
Poisson detector noise was injected to match reported signal-to-noise ratios.
To study model robustness, we applied Gaussian blurring and dose-specific shot
noise to simulate additional imaging imperfections.
Ground-truth atomic coordinates were rendered into Gaussian ``atom masks'' to
enable supervised training of localisation models.
Each sample thus forms an image--structure--property triplet:
(i) the noisy STEM projection,
(ii) the corresponding ground-truth \textsc{cif},
and (iii) DFT-level formation energy (along with band gap and magnetic moment,
if available).
We conducted principal component analysis (PCA) on structural fingerprints,
which revealed clear clustering patterns.
Subsequently, $k$-means clustering ($k{=}2,6$) was used to ensure balanced
train/validation/test splits across the chemical diversity of the dataset.

% --------- Two-column wide figure (spans both columns) ----------
\begin{figure*}[t]
  \centering
  \vspace{-0.2cm}
  \includegraphics[width=0.95\textwidth]{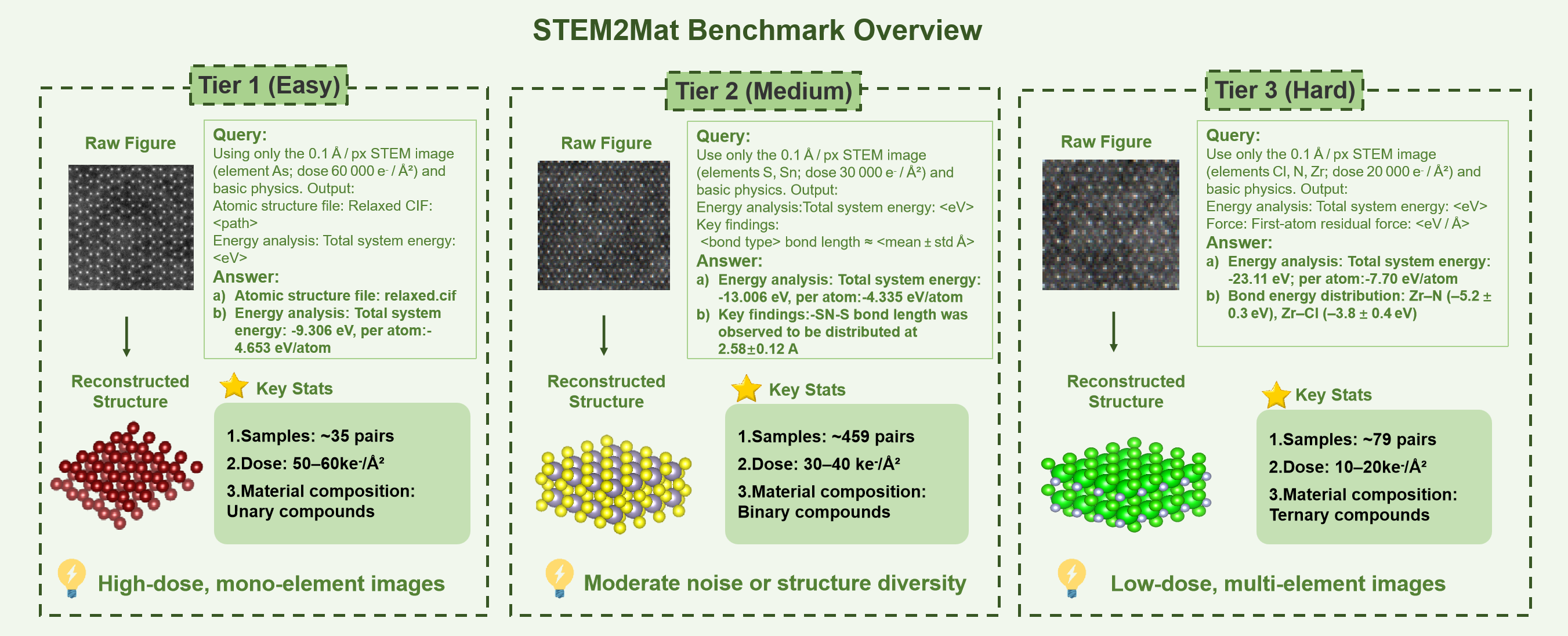}
  \caption{\textbf{Overview of the STEM2Mat benchmark design}, illustrating the
  tiered classification of STEM samples by material complexity and imaging
  dose, which systematically stratifies reconstruction difficulty from simple
  unary to complex ternary compounds.}
  \label{fig:benchmark_overview}
\vspace{-0.4cm}
\end{figure*}
% ---------------------------------------------------------------

\subsection{STEM2Mat Benchmark Split and Tiering}
To construct a representative and tractable benchmark dataset for STEM-based
crystal modeling, we applied stringent geometric and chemical screening
criteria to the 2,143 collected 2D material structures, ensuring their
suitability for monolayer imaging and end-to-end reconstruction.
Specifically, we retained only structures containing no more than three
distinct elements.
For those with multiple elements, we required a minimum atomic-number span of
ten, i.e., $\max(Z_i)-\min(Z_i)\ge 10$, to ensure sufficient imaging contrast
between heavy and light atoms.
To guarantee monolayer geometry, we limited the $z$-axis thickness to no more
than 3~\AA.
Each structure’s atomic coordinates were then projected onto the $(x,y)$ plane,
discretized onto a 1~\AA\ grid, and evaluated for overlapping projections.
Only structures with a projected duplication ratio below 10\%, i.e.,
$\frac{\text{Number of overlapping grid points}}{\text{Total grid points}}\le
0.1$, were retained to avoid ambiguity in atomic interpretation.

Following this multi-criteria filtering process, we retained exactly
\textbf{459 unique} well-defined, unambiguous monolayer structures---21\% of the
original dataset---for blind end-to-end evaluation.
The remaining 1,693 samples were split into training (80\%) and validation
(20\%) sets to support model training and tuning.

To analyze model performance as a function of task difficulty, we generated a total of
\textbf{570 test instances} from these 459 unique structures.
The tiers are stratified based on material composition and imaging noise
(see Fig.~\ref{fig:benchmark_overview}):

\textbullet\ \textbf{Tier 1} (35 instances) -- A subset of unary materials simulated under high electron
doses (\(5\text{--}6\times10^{4}\,\mathrm{e}^{-}/\si{\angstrom^{2}}\)); these
images are high contrast, low noise, and represent the easiest cases.

\textbullet\ \textbf{Tier 2} (459 instances) -- \textbf{The complete set of 459 unique structures} simulated under moderate electron dose
conditions (\(3\text{--}4\times10^{4}\,\mathrm{e}^{-}/\si{\angstrom^{2}}\));
this tier covers the full diversity of the test set with intermediate complexity.

\textbullet\ \textbf{Tier 3} (79 instances) -- A subset of ternary compounds or complex structures imaged at low dose
(\(1\text{--}2\times10^{4}\,\mathrm{e}^{-}/\si{\angstrom^{2}}\)); these samples
exhibit high noise, complex contrast, and are the most challenging to
reconstruct.

This design allows us to evaluate robustness against signal degradation on identical underlying topologies. With the STEM2Mat-Bench tiers defined, we now specify the quantitative metrics used throughout the paper.
For detailed definitions and formulae of the evaluation metrics, see
Section~\ref{sec:metrics}.

\subsection{Evaluation Metrics}
\label{sec:metrics}
To compare methods reproducibly, we report two \emph{primary} metrics---energy
error and lattice error---and two \emph{holistic} metrics that reflect chemical
and structural validity.

\paragraph{Mean Absolute Error (MAE).}
Average difference between predicted and DFT formation energies.
We note that Formation Energy MAE comprises two error sources: (i) geometric reconstruction error, and (ii) intrinsic MLIP prediction error.
Since the MLIP error is constant across all baselines (and bounded by the Oracle Upper Bound), the significant MAE reduction achieved by AutoMat is attributable to superior structural reconstruction accuracy.
\begin{equation}
  \mathrm{MAE}
  = \frac{1}{N}\sum_{i=1}^{N}
    \bigl|E^{\text{pred}}_{i}-E^{\text{ref}}_{i}\bigr|
  \;[\mathrm{meV/atom}]
  \label{eq:mae}
\end{equation}

\paragraph{Projected Lattice RMSD.}
Deviation of in-plane lattice constants:
\begin{equation}
  \mathrm{RMSD}_{xy} =
  \sqrt{\tfrac12\Bigl[(a^{\text{pred}}-a^{\text{ref}})^2+
                      (b^{\text{pred}}-b^{\text{ref}})^2\Bigr]}
  \label{eq:rmsd}
\end{equation}

\paragraph{Composition Correctness (C.C.).}
1 if elemental types and counts match the reference; 0 otherwise.

\paragraph{Structure Success Rate (S.S.).}
A prediction is successful when it satisfies both chemistry and geometry.
First define 2-D spatial similarity
\begin{equation}
  S_{\mathrm{spatial}}=\exp\!\bigl(-\mathrm{MSE}_{2\mathrm{D}}\bigr),
  \label{eq:spatial}
\end{equation}
where \(\mathrm{MSE}_{2\mathrm{D}}\) is the mean-squared error of projected
atomic positions after optimal element-wise matching.
Then
\begin{equation}
  \mathrm{S.S.}
  =\frac{1}{N}\sum_{i=1}^{N}
    \mathbb{I}\!\Bigl[S_{\mathrm{spatial}}^{(i)}\ge0.8\Bigr]\times100\%.
  \label{eq:ssr}
\end{equation}

% =========================== 4 Overview of AutoMat ===========================
\section{Overview of AutoMat}
\label{sec:overview_automat}

\subsection{Limitations of Existing Large Multimodal Models and Domain Tools}
\label{sec:agent_motivation}

Most current large multimodal models (e.g., GPT-4.1mini, LLama4V, Qwen2.5-VL,
ChemVLM) focus on general-purpose image understanding tasks such as scene
recognition, OCR, and molecular structure identification, but exhibit limited
capability in interpreting scientific images like electron microscopy (EM).
Domain-specific chemistry agents (e.g., ChemAgent, ChemCrow) remain
predominantly text-based, executing tool calls for predefined tasks without
closed-loop visual reasoning or image-guided decision-making. Specialized STEM
data toolkits (e.g., AtomAI) provide atomic coordinate extraction and
segmentation, yet their applicability is largely restricted to single-element
nanoparticles and fail with complex, multi-element STEM images. MicroscopyGPT,
the current state-of-the-art multimodal model in the domain, can generate
structural descriptions from user-provided STEM images but still relies on known
structural coordination and cannot directly reconstruct CIFs.

To date, existing agents are unable to autonomously generate simulation-ready
CIFs and predict formation energies from a single real STEM image, thus
completing a generalizable structure--property pipeline. To bridge this gap, we
introduce \textbf{AutoMat}, an agent framework with advanced reasoning, enabling
end-to-end modeling from pixel-level inputs to material property predictions.

% --------- Two-column wide figure (spans both columns) ----------
\begin{figure*}[t]
  \centering
  \vspace{-0.2cm}
  \includegraphics[width=0.95\textwidth]{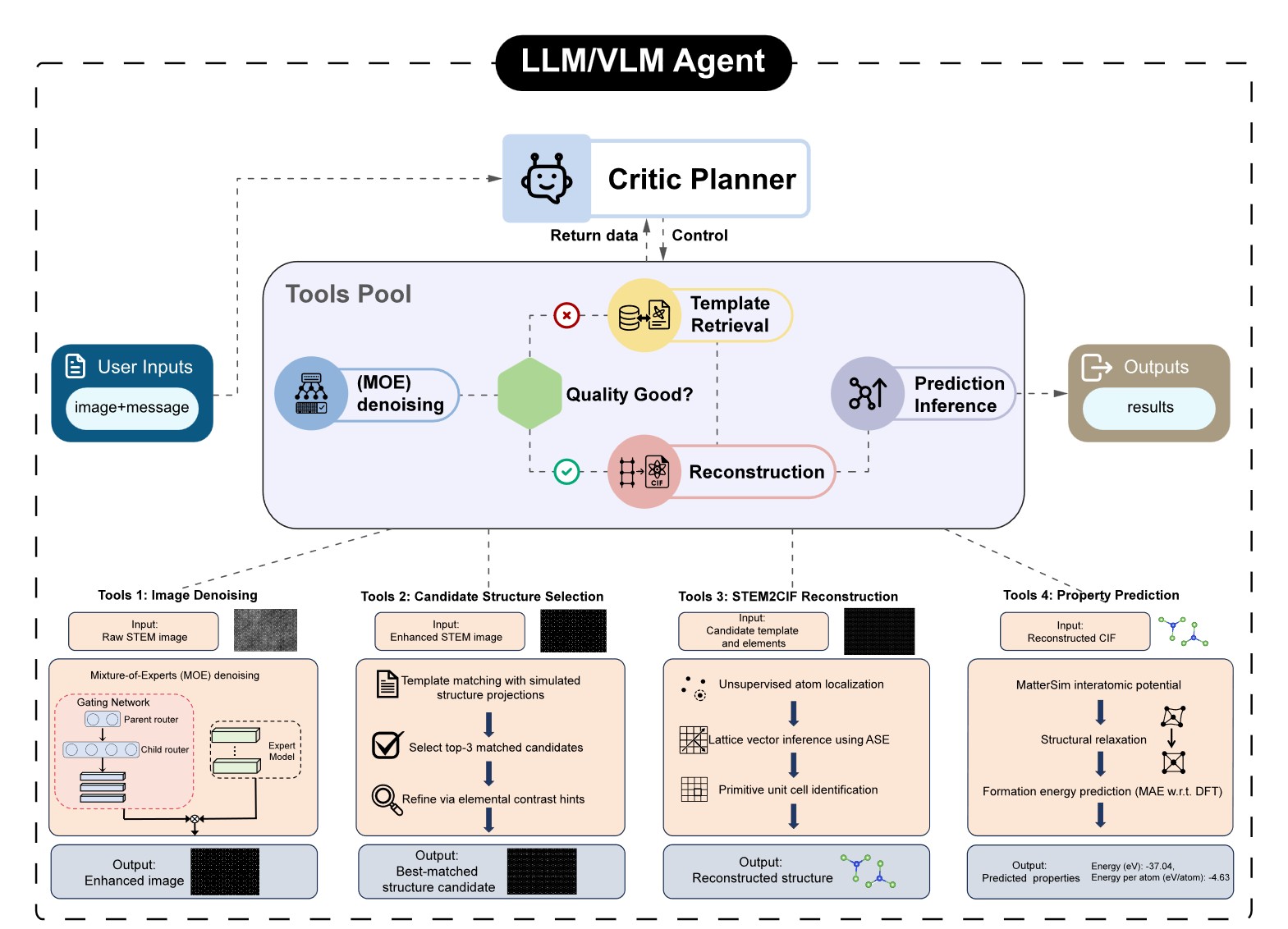}
  \caption{\textbf{AutoMat’s reasoning controller composes four stages}---denoising,
  template matching, structure reconstruction, and MLIP-based
  relaxation/property prediction---from STEM image to relaxed crystal with
  properties via closed-loop verification.}
  \label{fig:agent_pipeline}
\vspace{-0.4cm}
\end{figure*}
% ---------------------------------------------------------------

\subsection{Tools Pool}
Within AutoMat---a general and efficient framework---we employ four core tools:
MOE-DIVAESR, Image Template Matching, STEM2CIF, and MatterSim, responsible for
image denoising, template matching, structure reconstruction, and property
prediction, respectively. Further details are provided in
Appendices~\ref{appendix:moe_divaesr} and~\ref{appendix:template2cif}.

\textbf{MOE-DIVAESR} is a structure-pattern--adaptive mixture-of-experts (MoE)
model for STEM image denoising. Trained on an augmented STEM dataset, it
contains multiple expert networks, each specializing in a distinct structural
pattern, while a gating network dynamically selects the appropriate experts
based on the input image features. This design enables denoising, defect
correction, and fine-detail enhancement, producing sharply resolved atomic
columns.

\textbf{Image Template Matching} compares an enhanced image against a database
of known structures. By combining image features with elemental information,
this technique efficiently narrows down potential candidates and identifies the
structure that best fits experimental data. The process is even faster and more
reliable when researchers can directly match against pre-calculated or known
candidate structures.

\textbf{STEM2CIF} reconstructs a high-fidelity atomic model from the enhanced
image, optionally conditioned on a matched template when the retrieval branch
is invoked. It locates atomic column positions, infers lattice parameters,
and identifies atomic species. Crystallographic symmetry heuristics and physical 
constraints then reduce the model to its minimal repeating unit, yielding a 
crystallographic CIF file.

\textbf{MatterSim} is a pretrained machine-learning interatomic potential for
rapid structural relaxation and property evaluation. Trained on large-scale DFT
datasets, it attains near-DFT accuracy for energies and properties. Integrated
with ASE, it enables structure optimization and fast estimation of quantities
such as total energy, formation energy, and elastic moduli, providing a swift
and accurate alternative to conventional DFT.

\subsection{Flexible Tool-Calling Framework}
\paragraph{Formulation.}
We formulate reconstruction as a search process over a state space.
\textbf{State} $s$: current intermediate structure (e.g., denoised image, retrieved template, candidate CIF);
\textbf{Action} $a$: a tool execution (e.g., \texttt{Denoise}, \texttt{Retrieve}, \texttt{Refine});
\textbf{Verification} $\mathcal{V}(s)$: a physics-based critic (e.g., consistency checks via relaxation/simulation) that emits a failure signal when the current hypothesis is physically invalid;
\textbf{Policy} $\pi(a\mid s)$: a controller that triggers \textbf{rollback-and-retry} upon failure, effectively pruning invalid branches and performing inference-time search.

\paragraph{Physics-Guided Reasoning Loop.}
Building on this formulation, \textbf{AutoMat} establishes a decision control loop (Fig.~\ref{fig:agent_pipeline}) where the LLM backbone (e.g., \textbf{DeepSeekV3}) acts as a \emph{text-based reasoning controller}. Instead of processing pixels directly, it reasons over structured tool outputs \textbf{(e.g., quality metrics)} to re-plan the workflow dynamically.
In the execution phase, the policy $\pi$ adapts to the input state:
First, \textbf{MOE-DIVAESR} is called for pattern-adaptive denoising; if quality thresholds are not met, this step is repeated.
Next, the agent invokes \textbf{STEM2CIF} to localize atoms, fit lattices, and output confidence scores. \textbf{Image Template Matching} is not a mandatory first step; if the denoising quality is sufficient, the controller will skip retrieval. Retrieval serves as a \emph{state-dependent auxiliary branch}, triggered only when the direct path fails or a stronger structural prior is needed. Finally, \textbf{MatterSim} relaxes the structure and predicts properties for \textbf{physical validation}. By monitoring intermediate results and executing rollbacks upon failure, AutoMat achieves robust end-to-end automation without relying on any single model.

% =========================== 5 Experiments and Results ===========================
\section{Experiments and Results}
\label{sec:experiments}
\subsection{Quantitative Evaluation}

\paragraph{Baseline Overview.}
To make comparisons fair and interpretable, we group baselines by \emph{task
scope}:
\begin{itemize}[leftmargin=*, topsep=2pt, itemsep=1pt, parsep=0pt, partopsep=0pt]
  \item \textbf{Image $\rightarrow$ Property (no explicit structure).}
  GPT-4.1mini, Qwen2.5-VL (32B), LLama4V (17B), ChemVLM (8B) and MicroscopyGPT
  (11B) receive a fixed prompt, composition hints, and a STEM image to infer
  material properties. These baselines do \emph{not} produce simulation-ready
  CIFs, so we only compare them on energy metrics.
  \item \textbf{Image $\rightarrow$ Structure (structure-only tools).}
  AtomAI's segmentation network detects atomic centers; relative coordinates plus
  image resolution are used to fit the lattice. This baseline outputs atomic
  models and is evaluated on structural metrics.
  \item \textbf{Structure $\rightarrow$ Property (oracle bound).}
  The ground-truth CIF is fed directly to the MatterSim MLIP to benchmark
  formation-energy error under perfect-structure assumptions (MLIP-bias bound).
\end{itemize}

We summarize the performance of these baselines and our \textbf{AutoMat} system
on the test set in Tables~\ref{tab:energy-mae} and~\ref{tab:structure-metrics},
which together cover samples across Tier~1--3 difficulty levels. To further
separate tool-level capability from controller-level orchestration, we also evaluate same-tool scripted variants. In the main text, we report a stronger heuristic script with hand-coded quality gating, retrieval branching,
rollback, and bounded retries. Detailed ablations and backbone sensitivity
analyses are provided in Appendices~\ref{appendix:llm-vlm-comparison}
and~\ref{appendix:ablation}. Here we present results using DeepSeek (LLM) and
GPT-4o (VLM) as representative backbone models; experiments show that AutoMat
achieves comparable and consistent performance across different backbones.

%====================================================
% TABLE 1 : Formation-energy MAE (all methods)
%====================================================
\begin{table}[t]
\centering
\caption{\textbf{Formation-energy MAE (meV/atom) across tiers.}
Lower is better ($\downarrow$). ``--'' indicates the method does not provide an
energy prediction.}
\label{tab:energy-mae}
\small
\resizebox{\columnwidth}{!}{%
\begin{tabular}{lcccc}
\toprule
Method & Tier 1 $\downarrow$ & Tier 2 $\downarrow$ & Tier 3 $\downarrow$ & Avg. $\downarrow$ \\
\midrule
GPT-4.1 Vision       & 2521.9 & 2657.4 & 2817.7 & 2657.3 \\
Qwen-VL (32B)        & 3488.6 & 2673.9 & 4113.9 & 2763.5 \\
LLama4V (17B)        & 2549.8 & 2898.3 & 3662.6 & 2931.7 \\
ChemVLM (8B)         & 3961.1 & 4237.5 & 5090.3 & 4237.6 \\
MicroscopyGPT (11B)  & 590.51 & 616.23 & 680.28 & 620.22 \\
AtomAI               & --     & --     & --     & --     \\
\textbf{AutoMat (Ours, DeepSeek)} & \textbf{343.59} & \textbf{320.21} & \textbf{333.49} & \textbf{321.57} \\
\textbf{AutoMat (Ours, GPT-4o)}   & \textbf{341.72} & \textbf{322.05} & \textbf{334.12} & \textbf{323.25} \\
\midrule
Upper Bound          & 57.377 & 47.227 & 30.945 & 48.105 \\
\bottomrule
\end{tabular}%
}
\end{table}

%====================================================
% TABLE 2 : Structural metrics (only methods that output structures)
%====================================================
\begin{table}[t]
\centering
\caption{\textbf{Structural-accuracy metrics for methods that output atomic models.}
RMSD\textsubscript{xy}$\downarrow$: in-plane lattice RMSD (lower is better);
C.C.$\uparrow$: composition correctness; S.S.$\uparrow$: structure success rate.}
\label{tab:structure-metrics}
\small
\setlength{\tabcolsep}{3pt}
\resizebox{\columnwidth}{!}{%
\begin{tabular}{llccc}
\toprule
Tier & Method & RMSD\textsubscript{xy} (\AA)$\downarrow$ & C.C.\ (\%)$\uparrow$ & S.S.\ (\%)$\uparrow$ \\
\midrule
\multirow{4}{*}{\textbf{1}}
 & AtomAI        & 43.96$\pm$0.31 & 2.70  & 0.0 \\
 & MicroscopyGPT & 1.56$\pm$0.72  & 92.8  & 0.0 \\
 & \textbf{AutoMat (Ours, DeepSeek)} & \textbf{0.11$\pm$0.02} & \textbf{88.9} & \textbf{85.0} \\
 & \textbf{AutoMat (Ours, GPT-4o)}   & \textbf{0.11$\pm$0.02} & \textbf{89.2} & \textbf{85.3} \\
\midrule
\multirow{4}{*}{\textbf{2}}
 & AtomAI        & N/A           & 0.0   & 0.0 \\
 & MicroscopyGPT & 1.23$\pm$0.75 & 30.1  & 25.9 \\
 & \textbf{AutoMat (Ours, DeepSeek)} & \textbf{0.11$\pm$0.03} & \textbf{85.9} & \textbf{84.0} \\
 & \textbf{AutoMat (Ours, GPT-4o)}   & \textbf{0.12$\pm$0.02} & \textbf{86.4} & \textbf{84.7} \\
\midrule
\multirow{4}{*}{\textbf{3}}
 & AtomAI        & N/A           & 0.0   & 0.0 \\
 & MicroscopyGPT & 1.33$\pm$0.92 & 16.46 & 0.0 \\
 & \textbf{AutoMat (Ours, DeepSeek)} & \textbf{0.11$\pm$0.03} & \textbf{73.1} & \textbf{73.1} \\
 & \textbf{AutoMat (Ours, GPT-4o)}   & \textbf{0.11$\pm$0.02} & \textbf{72.8} & \textbf{72.8} \\
\bottomrule
\end{tabular}%
}
\end{table}

\begin{figure*}[t]
  \centering
  \includegraphics[width=0.95\textwidth]{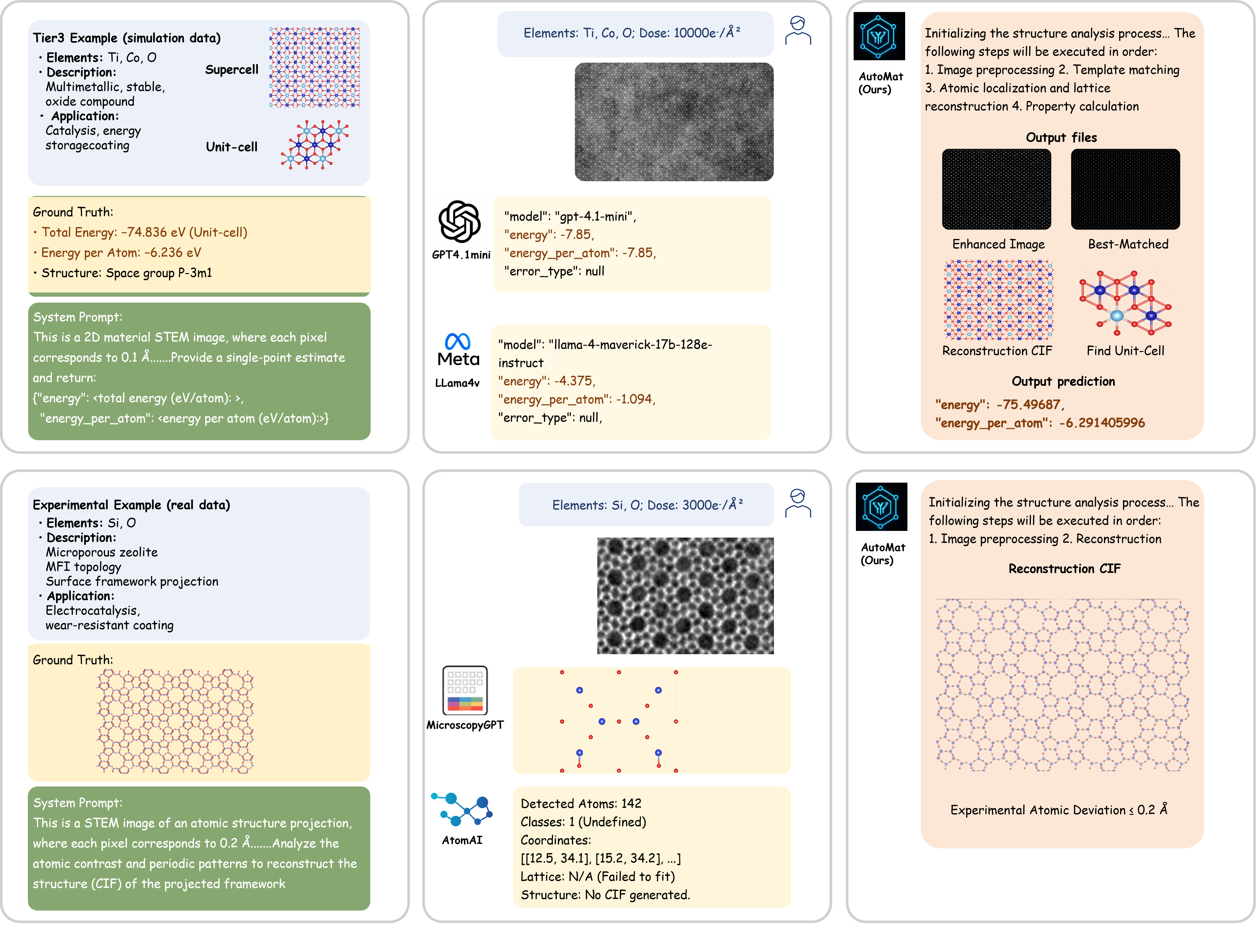}
  \caption{\textbf{Visualizing the reasoning gap.} We compare models on \textbf{(Top) a hard synthetic instance} and \textbf{(Bottom) a real-world ZSM-5 experiment}. While baselines suffer from \textit{geometric hallucinations} or lack of structure due to missing physical constraints, \textbf{AutoMat} provides initial evidence of zero-shot transfer by orchestrating a physics-guided reasoning loop to recover plausible surface topologies.}
  \label{fig:case_studies}
\end{figure*}

\begin{table}[t]
\centering
\caption{\textbf{Controller ablation against a stronger heuristic script.}
The heuristic script uses the same tools as AutoMat with hand-coded quality
gating, retrieval branching, rollback, and bounded retries. Energy MAE is
reported in meV/atom. Lower is better for Energy MAE; higher is better for
C.C. and S.S.}
\label{tab:controller_ablation_main}
\small
\setlength{\tabcolsep}{3pt}
\resizebox{\columnwidth}{!}{%
\begin{tabular}{llcccc}
\toprule
Metric & Method & Tier 1 & Tier 2 & Tier 3 & Overall \\
\midrule
\multirow{2}{*}{Energy MAE $\downarrow$}
 & Heuristic Script & 380.00 & 373.00 & 585.00 & 385.42 \\
 & \textbf{AutoMat} & \textbf{343.59} & \textbf{320.21} & \textbf{333.49} & \textbf{321.57} \\
\midrule
\multirow{2}{*}{C.C. (\%) $\uparrow$}
 & Heuristic Script & 81.0 & 77.2 & 63.4 & 76.5 \\
 & \textbf{AutoMat} & \textbf{88.9} & \textbf{85.9} & \textbf{73.1} & \textbf{83.2} \\
\midrule
\multirow{2}{*}{S.S. (\%) $\uparrow$}
 & Heuristic Script & 79.4 & 74.8 & 62.3 & 74.2 \\
 & \textbf{AutoMat} & \textbf{85.0} & \textbf{84.0} & \textbf{73.1} & \textbf{83.2} \\
\bottomrule
\end{tabular}%
}
\end{table}

\paragraph{Discussion of Results.}
For energy prediction, \textbf{AutoMat} achieves a mean formation energy MAE of
$332 \pm 12$~meV/atom, with tier-wise results of 343.59, 320.21, and
333.49~meV/atom. Although this is higher than the MLIP lower bound of
57~meV/atom, it is still far better than the multi-eV errors of
vision--language models. Even compared with MicroscopyGPT (used off-the-shelf
under a fixed prompt and without additional fine-tuning on STEM2Mat-Bench),
AutoMat remains much stronger, with MicroscopyGPT reaching only about
620~meV/atom. As task difficulty increases, most baselines show higher MAE,
confirming the soundness of our tiered benchmark. These results indicate that
AutoMat’s remaining errors come mainly from reconstruction rather than MLIP
limits, and that its predicted structures are reliable for downstream property
evaluation. Figure~\ref{fig:case_studies} visualizes the reasoning gap: the top panel (Tier~3) shows that while baselines suffer from geometric hallucinations under low-dose conditions, AutoMat accurately recovers the lattice. The bottom panel demonstrates \textbf{zero-shot generalization} to real experimental ZSM-5 images, verifying robustness beyond the synthetic training distribution.

For structural reconstruction, AutoMat achieves an average projected
RMSD$_{xy}$ of about 0.11~\AA{}, much lower than MicroscopyGPT
(1.3--1.6~\AA{}, requiring prior coordination knowledge) and AtomAI
(43--44~\AA{}). Most deviations can be corrected through final relaxation. In
composition correctness, AutoMat averages 83\% across tiers (88.9\%, 85.9\%,
73.1\%), while MicroscopyGPT performs well only in Tier~1 ($\approx$92\%) but
drops sharply in Tier~2--3. AtomAI, by contrast, stays below 2.7\%. For
structure success rate, AutoMat achieves 83.2\% overall (85.0\%, 84.0\%,
73.1\%), compared with only 25.9\% for MicroscopyGPT in Tier~2 and almost zero
for AtomAI.

Beyond comparisons with external baselines, we further compare AutoMat with a
stronger same-tool heuristic script to test whether the controller contributes
beyond a well-engineered non-LLM pipeline. This heuristic script already uses
the same scientific tools as AutoMat and includes hand-coded quality gating,
retrieval branching, rollback, and bounded retries. As shown in
Table~\ref{tab:controller_ablation_main}, it provides a strong scripted
baseline but remains consistently below AutoMat across energy and structural
metrics. The gap is especially clear on Tier~3, where energy MAE decreases from
585.00 to 333.49 meV/atom and S.S. improves from 62.3\% to 73.1\%. Overall,
AutoMat also improves Energy MAE from 385.42 to 321.57 meV/atom, C.C. from
76.5\% to 83.2\%, and S.S. from 74.2\% to 83.2\%. These results suggest that
the gain of AutoMat is not merely due to the underlying tools or a few threshold
rules, but to more flexible failure-aware orchestration and parameter adjustment
under noisy, ambiguous, and multi-element reconstruction conditions.

In summary, \textbf{AutoMat} not only outperforms all baselines but also
maintains strong results in challenging Tier~3 cases with multi-element
compositions and low imaging doses, demonstrating robustness and
generalizability across the full benchmark.

% =========================== Real Case + Human Effort ===========================
\subsection{Real iDPC-STEM Case Study and Human-Expert Effort}
\noindent\textbf{Real iDPC-STEM case.} 
To complement the simulation-based benchmark, we also evaluate \textbf{AutoMat} on a real iDPC-STEM case: a ZSM-5 zeolite sample acquired on the same Cs-corrected STEM under dose-limited conditions. Without any manual tuning, the agent automatically performs denoising and structural reconstruction, and outputs a CIF structure whose projected lattice and channel framework qualitatively agree with the known crystallographic model. As shown in Figure~\ref{fig:case_studies} (bottom) and Appendix~\ref{appendix:real_stem_zsm5}, AutoMat provides preliminary evidence of transfer from 2D monolayer simulations to the 3D surface projection of the zeolite. This demonstrates that AutoMat has a certain ability to transfer the image--structure--property pipeline from abTEM simulations to real microscope images.

\noindent\textbf{Human-expert effort.}
For comparison, we consulted a senior electron-microscopy expert in our group.
For Tier-2-like STEM images, manually interpreting a 1024$\times$1024 image,
confirming lattice parameters and elemental species, and producing a
simulation-ready CIF typically requires about 6--8 hours per sample. In
contrast, on our benchmark hardware, \textbf{AutoMat} processes similar Tier-2
samples in roughly 2 minutes per case, yielding orders-of-magnitude higher
throughput.

% ================================ Error Analysis ==============================
\subsection{Error Analysis}
To better understand the failure modes of \textbf{AutoMat}, we analyzed
representative failure cases across all three tiers and identified two primary
types of errors (a detailed analysis with examples is provided in
Appendix~\ref{appendix:failure_analysis}):

\textbf{(1) Template retrieval failure (39.3\%).}
In these cases, AutoMat failed to retrieve the correct structure from the
template database, resulting in mismatches in atomic arrangements and element
types. This led to cascading errors in structure, composition, and property
predictions. Incorrect atom counts further caused large energy estimation
errors.

\textbf{(2) Downstream failure despite correct template (60.7\%).}
Even with the correct template, downstream steps failed due to projection
ambiguity or elemental confusion. In 40\% of these cases, atoms appeared too
close in the 2D projection, and the lack of z-axis information led to poor
relaxation and inaccurate energy estimates. In 20.7\%, elements with similar
atomic numbers (e.g., C and O) exhibited indistinguishable contrast, causing
misclassification and full breakdowns in lattice fitting and CIF generation.

These findings highlight two key directions for improving \textbf{AutoMat}:
(i) improving retrieval robustness via uncertainty-aware methods; and (ii)
overcoming 2D projection limits through 3D-aware modeling
\citep{guo2023viewrefer,tang2024any2point,tang2025exploring} and enhanced modality
integration. To make projection ambiguity visible to users, \texttt{STEM2CIF}
also reports per-site confidence scores based on local fitting quality and
nearest-neighbor geometry, and appends warning flags for low-confidence regions
(see Appendix~\ref{appendix:template2cif}). In addition, adaptive distribution
alignment and graph-domain adaptation may help reduce the remaining gap between
simulated STEM images and real experimental inputs~\citep{chen2026learning}.
To further distinguish template-free reconstruction from unseen-family
generalization, we provide a leave-one-chemical-family-out evaluation in
Appendix~\ref{appendix:ablation}.

% ================================= Conclusion =================================
\section{Conclusion}
\label{sec:conclusion}
We proposed \textbf{AutoMat}, an end-to-end agent system that
reconstructs material structures and predicts properties from STEM images. By
integrating pattern-adaptive vision models, symmetry-constrained reconstruction,
and LLM-driven orchestration, AutoMat achieves accurate alignment between
microscopy data and atomic modeling, significantly outperforming existing
methods in structural and energetic evaluation. Meanwhile, we propose
STEM2Mat-Bench for this task. AutoMat advances autonomous materials research and
AI-driven experimentation. Future work will focus on strengthening its role as a
bridge between experimental characterization and theoretical computation.

\section*{Acknowledgements}

This work was supported by the National Natural Science Foundation of China
(Nos. 22322203 and 22275110), the Tsinghua University Initiative Scientific
Research Program, the Key Research and Development Program of Inner Mongolia
and Ordos (Nos. Ordoslab-kjzc-202506 and Ordoslab-sysjc-202502), the HE
Science Foundation, the Scientific Research Innovation Capability Support
Project for Young Faculty (No. ZYGXQNJSKYCXNLZCXM-E7), and the Central
Universities Young Faculty Research Innovation Capacity Support Program
(No. SRICSPYF-ZY2025039).

\section*{Impact Statement}

This paper presents \textit{AutoMat}, an agent that automates microscopy-to-structure reconstruction from iDPC-STEM images and enables downstream materials evaluation. The expected positive impact is to reduce manual workload and accelerate reproducible materials analysis.

Potential risks include incorrect reconstructions (e.g., under low-dose noise, similar-$Z$ elements, or projection ambiguity) being propagated into downstream simulations and property predictions, which could mislead scientific conclusions if used without verification. We therefore position AutoMat as an assistive tool and recommend inspecting intermediate outputs (retrieved templates, reconstructed structures, and logs) and applying expert validation, especially for out-of-distribution experimental conditions. AutoMat should not be used as a fully autonomous replacement for expert crystallographic analysis in high-stakes experimental studies; rather, it is intended to assist experts by accelerating candidate generation, consistency checking, and downstream validation.


\bibliography{icml2026}
\bibliographystyle{icml2026}

%%%%%%%%%%%%%%%%%%%%%%%%%%%%%%%%%%%%%%%%%%%%%%%%%%%%%%%%%%%%%%%%%%%%%%%%%%%%%%%
%%%%%%%%%%%%%%%%%%%%%%%%%%%%%%%%%%%%%%%%%%%%%%%%%%%%%%%%%%%%%%%%%%%%%%%%%%%%%%%
% APPENDIX
%%%%%%%%%%%%%%%%%%%%%%%%%%%%%%%%%%%%%%%%%%%%%%%%%%%%%%%%%%%%%%%%%%%%%%%%%%%%%%%
%%%%%%%%%%%%%%%%%%%%%%%%%%%%%%%%%%%%%%%%%%%%%%%%%%%%%%%%%%%%%%%%%%%%%%%%%%%%%%%
\newpage

\appendix
% 如果您想要整个附录变成单栏（推荐），取消下面这行的注释即可：
% \onecolumn 

\section{Appendix}
\label{sec:appendix}

% ------------------------- Dataset Overview -------------------------
\subsection{Dataset Overview}
\label{appendix:dataset_overview}
\begin{figure*}[t]
  \centering
  \includegraphics[width=0.92\linewidth]{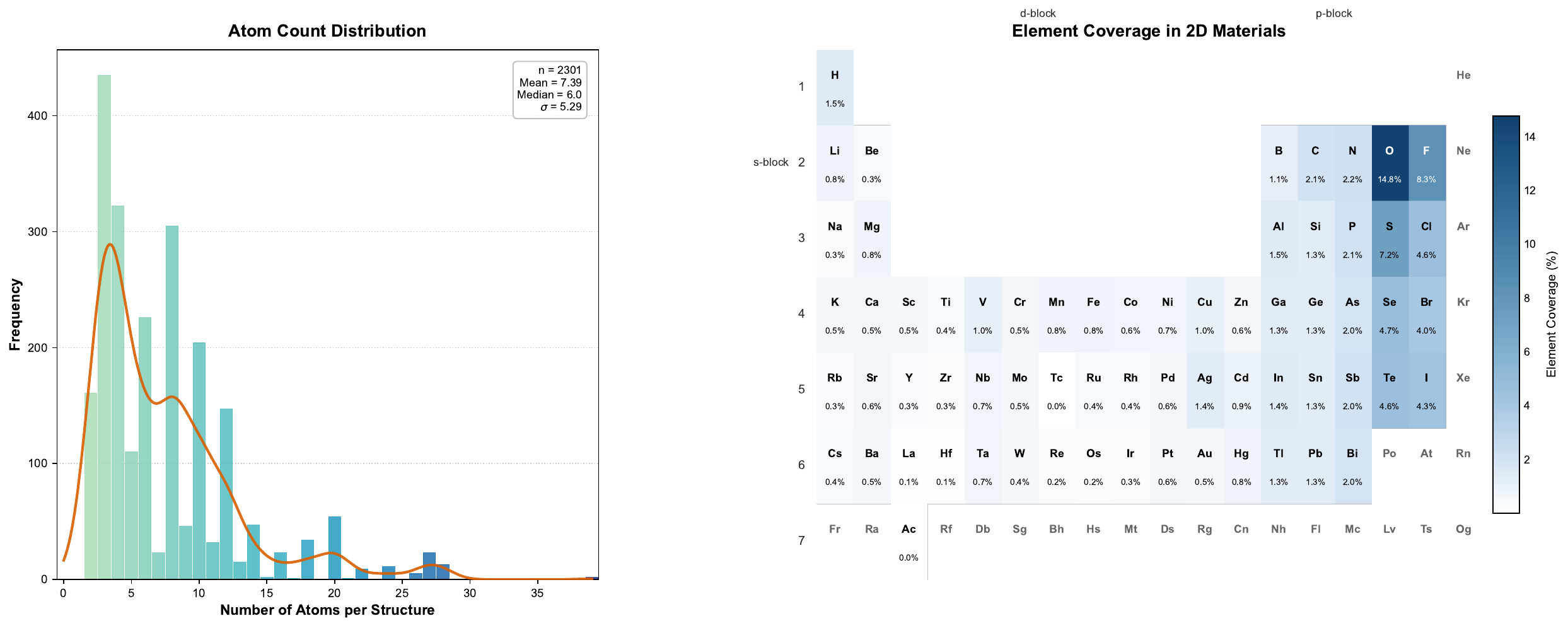}
  \caption{Distribution of per-unit-cell structure atom counts (left) and elemental coverage percentages in our curated 2D materials dataset (right).}
  \label{fig:data_overview_appendix}
\end{figure*}

As illustrated in Figure~\ref{fig:data_overview_appendix}, the atom count distribution (left) reveals a predominance of compact unit cells (median $N=6.0$), ensuring computational feasibility for structural reasoning. The periodic table heatmap (right) demonstrates comprehensive elemental coverage spanning 67 unique species, with intensity peaks in chalcogens (e.g., O, S) and transition metals that accurately reflect the compositional diversity of stable 2D materials.

%============================= Appendix Methods =============================
\subsection{MOE--DIVAESR: Training Pipeline \& Architecture}
\label{appendix:moe_divaesr}

\noindent\textbf{A.\ Training pipeline.}
\begin{enumerate}[leftmargin=2.2em, itemsep=2pt, topsep=2pt]
  \item \textbf{Dataset partitioning.}
  A total of 1,693 denoised STEM images were projected to a PCA-reduced feature space and clustered using \textit{k}-means ($k{=}8$).
  The resulting clusters were grouped into \textbf{2 parent} and \textbf{6 child} routing categories, defining the hierarchical MoE.

  \item \textbf{Prototype selection \& simulation.}
  For each child cluster, the top 25\% most representative samples (424 total) were re-simulated into iDPC-STEM projections.
  These images were augmented with random rotation, flipping, Poisson noise, Gaussian blur, elastic deformation, and sliding-window cropping,
  resulting in a total of \textbf{756,042} training patches of size $128 \times 128$.

  \item \textbf{Stage 1: Router pretraining.}
  A ResNet-18 model was trained as a routing network for 30 epochs to predict the appropriate expert index.
  It achieved a top-1 routing accuracy of \textbf{99\%} on the validation set. The weights were then frozen for expert training.

  \item \textbf{Stage 2: Expert training.}
  Six DIVAESR experts (EDSR backbone with 12 residual blocks and 64 feature maps) were trained jointly under fixed routing for 300 epochs.
  Compared to a single DIVAESR baseline, the MoE model reduced reconstruction error by approximately 50\%.
\end{enumerate}

\medskip
\noindent\textbf{B.\ Model Architecture and Objective.}
\smallskip

Each expert $E_k$ is modeled as a two-stage cascade of a denoising module and a super-resolution module:
\begin{equation}
  E_k = g_{\phi_k}^{\text{SR}} \circ f_{\theta_k}^{\text{DIVAE}},
  \qquad k = 1, \dots, K .
  \label{eq:moe_expert}
\end{equation}
where $f_{\theta_k}^{\text{DIVAE}}$ is the denoising network and $g_{\phi_k}^{\text{SR}}$ is the super-resolution network.

The router $R_{\psi}$ produces a one-hot routing mask $\mathbf{r} \in \{0,1\}^K$ that selects expert $k^*$ for each input:
\begin{equation}
  k^* = R_{\psi}\!\left(x^{\text{noisy}}\right).
  \label{eq:router_selection}
\end{equation}

The output of the MoE model is then computed as:
\begin{equation}
  \hat{x}^{\text{HR}} = \sum_{k=1}^{K} r_k \cdot
  g_{\phi_k}^{\text{SR}} \left( f_{\theta_k}^{\text{DIVAE}}(x^{\text{noisy}}) \right).
  \label{eq:moe_output}
\end{equation}

\paragraph{Joint loss objective.}
The total training loss for a mini-batch is given by:
\begin{equation}
  \mathcal{L}_{\text{MoE}} = \sum_{k=1}^{K} r_k \cdot
  \left( \mathcal{L}_{\text{DIVAE}}^{(k)} + \mathcal{L}_{\text{SR}}^{(k)} \right)
  + \lambda \sum_{k=1}^{K} \bar{p}_k \log \bar{p}_k .
  \label{eq:moe_loss}
\end{equation}
where
\begin{equation}
  \mathcal{L}_{\text{DIVAE}}^{(k)} = \frac{1}{N} \sum_{i=1}^{N}
  \left\| \hat{x}_i^{\text{clean}} - x_i^{\text{clean}} \right\|_2^2
  + \beta \cdot \mathrm{KL} \left[ q(z|x_i^{\text{noisy}}) \,\|\, \mathcal{N}(0, I) \right]
  \label{eq:ldivae}
\end{equation}
is the $\beta$-VAE denoising loss, and
\begin{equation}
  \mathcal{L}_{\text{SR}}^{(k)} = \frac{1}{N} \sum_{i=1}^{N}
  \left\| \hat{x}_i^{\text{HR}} - x_i^{\text{HR}} \right\|_1
  \label{eq:lsr}
\end{equation}
is the L1 reconstruction loss for super-resolution.
Here, $\bar{p}_k$ denotes the empirical routing frequency for expert $k$.

This formulation jointly minimizes noise and detail loss through specialized expert modules.
The KL regularization in Eq.~\eqref{eq:ldivae} stabilizes latent representations, while Eq.~\eqref{eq:lsr} preserves fine image details.
The entropy term in Eq.~\eqref{eq:moe_loss} promotes balanced expert usage.

% -------------------- Template Matching and STEM2CIF --------------------
\subsection{Template Matching and STEM2CIF Conversion}
\label{appendix:template2cif}

\noindent\textbf{A.\ Image Template Matching.}
\begin{enumerate}[leftmargin=2.2em, itemsep=2pt, topsep=2pt]
  \item \textbf{Feature extraction.}
  Atomic centroids $\{\mathbf{p}_i\}_{i=1}^{N}$ are detected in each denoised image.
  An $m{=}32$-bin radial-distribution histogram $\mathbf{g}=(g_1,\dots,g_m)$ is computed:
  \begin{equation}
    g_b = \frac{1}{N} \sum_{i<j}
    \mathbb{I}\!\left[r_{b-1}\le \|\mathbf{p}_i-\mathbf{p}_j\| < r_b\right],
    \quad b=1,\dots,m .
    \label{eq:rdf_appendix}
  \end{equation}

  \item \textbf{Candidate retrieval.}
  All template images are similarly encoded as RDF descriptors $\{\mathbf{g}^{(t)}\}$.
  The cosine distance is computed as:
  \begin{equation}
    d_{\cos}\left(\mathbf{g},\mathbf{g}^{(t)}\right)
    = 1 - \frac{\langle \mathbf{g}, \mathbf{g}^{(t)} \rangle}
    {\|\mathbf{g}\| \cdot \|\mathbf{g}^{(t)}\|}.
    \label{eq:cosine_appendix}
  \end{equation}
  The top-$k$ nearest templates ($k=10$ by default) are retrieved.

  \item \textbf{Element filtering.}
  Retain only templates whose elemental composition $\mathcal{E}^{(t)}$ matches the input:
  \begin{equation}
    \mathcal{E}^{(t)} = \mathcal{E}^{(\text{input})}.
  \end{equation}
  From the remaining candidates, select the one with the smallest $d_{\cos}$.

  \item \textbf{Output.}
  The top-1 matched template is copied to the output directory for downstream reconstruction.
\end{enumerate}

\medskip
\noindent\textbf{B.\ STEM2CIF Conversion.}
\begin{enumerate}[leftmargin=2.2em, itemsep=2pt, topsep=2pt]
  \item \textbf{Peak localization.}
  Atomic column centers are localized using weighted mean-shift clustering followed by sub-pixel 2D Gaussian fitting.

  \item \textbf{Lattice fitting.}
  An initial lattice is fitted using least-squares optimization based on known pixel resolution (0.1~\AA/pixel), constrained by the space-group symmetry of the matched template.

  \item \textbf{Element assignment.}
  Atomic column intensities are compared to simulated iDPC-STEM contrast values, and the most probable element is assigned to each atomic site.

  \item \textbf{CIF generation.}
  The reconstructed periodic structure is reduced to its primitive unit cell and exported as a standard CIF file for downstream MatterSim relaxation and property prediction.
\end{enumerate}

\medskip
\noindent\textbf{C.\ Confidence Estimation and Warning Flags.}

To expose projection ambiguity to users, \texttt{STEM2CIF} additionally reports
a per-site confidence score during reconstruction. The confidence is estimated
from local lattice fitting quality and neighborhood geometry, using the local
fitting residual, represented by the fitted $R^2$, together with the nearest
neighbor (NN) distance. Intuitively, severely overlapping atomic columns or
closely packed heavy atoms tend to produce lower confidence, while isolated
well-fitted atoms yield higher confidence. When the confidence score falls below
a predefined safe threshold, AutoMat appends a warning flag to the reconstructed
output, indicating that the corresponding region should be inspected by an
expert before downstream property analysis.

\begin{table}[h]
  \centering
  \caption{\textbf{Examples of per-site confidence estimates from STEM2CIF.}
  Closely packed heavy atoms exhibit lower confidence, while well-separated
  atoms yield higher confidence.}
  \label{tab:confidence_estimation}
  \small
  \setlength{\tabcolsep}{4pt}
  \resizebox{\columnwidth}{!}{%
  \begin{tabular}{ccccccc}
    \toprule
    ID & Element & $x$ (\AA) & $y$ (\AA) & $R^2$ & NN Dist. (\AA) & Confidence \\
    \midrule
    65 & I & 12.85 & 65.10 & 0.972 & 1.48 & 0.744 \\
    53 & I & 13.29 & 62.46 & 0.969 & 1.67 & 0.654 \\
    21 & B & 40.54 & 35.65 & 0.977 & 6.22 & 0.953 \\
    41 & B & 68.28 & 53.80 & 0.993 & 2.91 & 0.876 \\
    \bottomrule
  \end{tabular}%
  }
\end{table}

As shown in Table~\ref{tab:confidence_estimation}, closely packed iodine
columns receive lower confidence scores, whereas more isolated boron sites
receive higher confidence scores. This mechanism does not fully resolve the
single-projection depth ambiguity, but converts part of the physical ambiguity
into a user-facing uncertainty signal.

% ------------------------------- Ablation --------------------------------
\subsection{Ablation Study}
\label{appendix:ablation}

The \texttt{STEM2CIF} module is the only component that converts microscopy images into atomic structures in the form of \texttt{CIF} files.
Removing it would prevent structural outputs, making RMSD, composition correctness, and formation-energy MAE undefined.
Therefore, ablating \texttt{STEM2CIF} is not meaningful.

To isolate contributions of other modules, we selectively disable
(i) pattern-adaptive denoising (MOE-DIVAESR), and
(ii) physics-guided template matching, while keeping the rest unchanged.
We report formation-energy MAE in Table~\ref{tab:ablation}.

% 【修改点2】小表格保持在单栏，使用 [h] 或 [t]
\begin{table}[h]
  \centering
  \caption{Energy per-atom MAE (meV/atom) under different ablation settings. Lower is better.}
  \label{tab:ablation}
  \small
  \begin{tabular}{lccc}
    \toprule
    \textbf{Method} & \textbf{Tier 1} & \textbf{Tier 2} & \textbf{Tier 3} \\
    \midrule
    \textit{No Denoising} & 6584 & 2616 & 938 \\
    \textit{No Template Retrieval}      & 617  & 608  & 672 \\
    \textbf{AutoMat (Full)}        & \textbf{344} & \textbf{320} & \textbf{333} \\
    \bottomrule
  \end{tabular}
\end{table}

Removing MOE-DIVAESR significantly increases error in Tier~1, indicating its critical role in enhancing image quality.
Removing template matching impacts Tier~3 most, where strong priors are essential under low-dose and multi-element conditions.
These results suggest that template retrieval should not be interpreted as a
purely marginal fallback. Rather, the current system is better viewed as
combining two complementary reconstruction paths: a direct image-driven path
via STEM2CIF, and a retrieval-guided path that provides a stronger structural
prior when direct reconstruction becomes unreliable. The ``No Template Retrieval''
setting shows that the direct path remains operative, but the sharp degradation
from 333 to 672 meV/atom on Tier~3 indicates that retrieval is especially
important in hard low-dose, multi-element cases. We therefore revise the main
text to describe retrieval as a \emph{state-dependent auxiliary branch} rather
than a purely secondary add-on.

\paragraph{Leave-One-Chemical-Family-Out (LOFO) Evaluation.}
To directly assess generalization beyond seen structural families, we conduct a
leave-one-chemical-family-out (LOFO) evaluation. For each held-out family, we
remove that family from both the training split and the retrieval library, and
test only on that family. We report three settings: \textbf{Full AutoMat},
\textbf{LOFO-Full}, and \textbf{LOFO-NoTemplate}. The first measures the
original in-family performance, the second measures unseen-family
generalization under the full controller, and the third measures the standalone
performance of the direct STEM2CIF path when no correct retrieval prior is
available.

\begin{table}[h]
  \centering
  \caption{Leave-one-chemical-family-out (LOFO) evaluation. Lower is better for RMSD$_{xy}$ and Energy MAE; higher is better for S.S. and C.C.}
  \label{tab:lofo}
  \small
  \setlength{\tabcolsep}{4pt}
  \resizebox{\columnwidth}{!}{
  \begin{tabular}{llcccc}
    \toprule
    \textbf{Held-out family} & \textbf{Setting} & \textbf{RMSD$_{xy}$ (\AA)} & \textbf{S.S. (\%)} & \textbf{C.C. (\%)} & \textbf{Energy MAE} \\
    & & $\downarrow$ & $\uparrow$ & $\uparrow$ & \textbf{(meV/atom)} $\downarrow$ \\
    \midrule
    \multirow{3}{*}{MXenes ($N{=}6$)}
      & Full AutoMat      & 0.117 & 100.0 & 100.0 & 174 \\
      & LOFO-Full         & 0.122 & 100.0 & 97.2  & 186 \\
      & LOFO-NoTemplate   & 0.181 & 83.3  & 83.3  & 345 \\
    \midrule
    \multirow{3}{*}{Perovskites ($N{=}53$)}
      & Full AutoMat      & 0.124 & 76.7  & 82.1  & 356 \\
      & LOFO-Full         & 0.125 & 66.8  & 73.6  & 370 \\
      & LOFO-NoTemplate   & 0.183 & 54.9  & 60.4  & 685 \\
    \bottomrule
  \end{tabular}}
\end{table}

The LOFO results help separate two questions that were previously conflated.
First, \textbf{LOFO-Full} measures unseen-family generalization under the full
controller. On MXenes, the drop from Full AutoMat to LOFO-Full is minimal,
while on perovskites the degradation is moderate rather than catastrophic,
indicating that AutoMat retains meaningful generalization beyond seen families.
Second, \textbf{LOFO-NoTemplate} measures how well the direct STEM2CIF path
works when no correct retrieval prior is available. Here the performance drops
much more substantially, especially for perovskites (356 $\rightarrow$ 685
meV/atom), showing that the direct path remains operative but that the
retrieval-guided branch is particularly important in structurally harder
families. Overall, these results suggest that AutoMat is neither a pure
retrieval system nor a fully retrieval-free one: it is better viewed as a
dual-path framework combining direct reconstruction with a state-dependent
retrieval-guided prior branch.

% ---------------------------- Broader Impact ----------------------------
\subsection{Broader Impact}
\label{appendix:broader_impact}

AutoMat can accelerate the discovery and validation of novel materials by automating structure reconstruction from electron microscopy images.
This reduces reliance on expert annotation and lowers the barrier to entry in under-resourced research settings.
Open release of data and code promotes transparency and reproducibility, fostering collaboration.

\subsection{Code and Data Availability}
\label{appendix:code_data_availability}

The code and dataset are publicly available at
\url{https://github.com/yyt-2378/AutoMat}
and
\url{https://huggingface.co/datasets/yaotianvector/STEM2Mat}.
The released repository includes the main AutoMat pipeline, evaluation scripts,
and instructions for reproducing the benchmark experiments.
% ------------------------- Experimental Settings -------------------------
\subsection{Experimental Settings}
\label{appendix:exp-settings}

\paragraph{Framework.}
Training and experiment management use \texttt{PyTorch Lightning} on a $4\times$ NVIDIA H100 NVLink server (80GB per GPU).
Hyperparameter tuning for MOE-DIVAESR required approximately 4--5 days (see Table~\ref{tab:moe_divaesr_config}).

% 长表格在单列如果放不下，可以调整字号
\begin{table}[h]
\centering
\caption{Configuration of MOE-DIVAESR and related submodules.}
\label{tab:moe_divaesr_config}
\footnotesize % 缩小字号防止溢出
\renewcommand{\arraystretch}{1.1}
\begin{tabularx}{\linewidth}{@{}p{0.55\linewidth}X@{}}
\toprule
\multicolumn{2}{c}{\textbf{Gating Network}} \\
\midrule
Parent Router Modules & 2 \\
Child Router Modules  & 6 \\
\midrule
\multicolumn{2}{c}{\textbf{Latent Prior Module (DIVAESR)}} \\
\midrule
Input Channels   & 1 \\
Latent Dimension & 128 \\
\midrule
\multicolumn{2}{c}{\textbf{Super-Resolution Module (DIVAESR)}} \\
\midrule
Residual Blocks        & 12 \\
Feature Maps per Block & 64 \\
Activation Function    & ReLU \\
Patch Size             & 128 \\
Scale Factor           & $\times2$ \\
Image Channels         & 1 (grayscale) \\
Precision              & FP32 \\
\midrule
\multicolumn{2}{c}{\textbf{Optimization and Training Settings}} \\
\midrule
Optimizer              & Adam \\
Initial Learning Rate  & 0.001 \\
Weight Decay           & 0.0 \\
KL Divergence Weight   & 0.00025 \\
Training Epochs        & 300 \\
Training Batch Size    & 128 \\
\bottomrule
\end{tabularx}
\end{table}

% ------------------------ MicroscopyGPT Baseline ------------------------
\subsection{MicroscopyGPT Baseline}
\label{appendix:microscopygpt}

MicroscopyGPT is used as an off-the-shelf vision--language baseline without additional fine-tuning.
Following Choudhary \textit{et al.}, MicroscopyGPT fine-tunes an 11B LLaMA-3.2--Vision model with QLoRA on simulated STEM images.
Each example pairs a 2D STEM projection (mainly along $(001)$) with a structured textual description containing lattice parameters, lattice angles, element types, and fractional atomic coordinates.

We do \emph{not} further fine-tune MicroscopyGPT on STEM2Mat-Bench to avoid overfitting and to keep baselines comparable.
For each benchmark image, we (i) feed the STEM image with a standard structure-generation prompt, (ii) parse the textual output into a structured representation, and (iii) convert it into a CIF file using ASE utilities.

% --------------- Robustness Across LLM/VLM Agents + Retry ---------------
\subsection{Robustness Across Agents and Retry}
\label{appendix:llm-vlm-comparison}

\paragraph{Retry Mechanism.}
We implement \textbf{in-context learning from environmental feedback}, where tool error messages are fed back into subsequent turns.
This enables the agent to refine calls until a maximum retry limit is reached or the tool call succeeds. 
This retry-and-repair mechanism is motivated by the broader need to improve LLM resilience under erroneous reasoning and perturbed execution states.

{\footnotesize
\begin{verbatim}
[TOOL_CALL] reconstruction_tool
{"img_path": ..., 
"elements": ["Co", "Fe", "O"]}}

[TOOL_RESPONSE] reconstruction_tool
ValueError: Unexpected "}"

[ANSWER]
Malformed arguments detected. 
Correcting and retrying.

[TOOL_CALL] reconstruction_tool
{"img_path": ..., 
"elements": ["Co", "Fe", "O"]}

[TOOL_RESPONSE] reconstruction_tool
{success: true, cif_path: ...}

[TOOL_CALL] validation_tool
{"cif_path": ...}

[TOOL_RESPONSE] validation_tool
{converged: true, stress_xx_GPa: -10.956}

[ANSWER]
High residual stress detected. 
Returning a low-confidence warning.
\end{verbatim}
}

\paragraph{State-Dependent Tool Invocation.}
AutoMat does not execute tools in a fixed order. Instead, tool invocation is conditioned on intermediate states such as denoising quality and reconstruction validity. In particular, template retrieval is not always required: when the denoised image is sufficiently reliable, the agent can directly invoke STEM2CIF and skip retrieval; retrieval is only triggered when the direct path fails or is judged unreliable.

{\footnotesize
\begin{verbatim}
[TOOL_CALL] denoise_tool
{"image_path": ...}

[TOOL_RESPONSE] denoise_tool
{success: true, recon_img: ...}

[ANSWER]
Denoising completed. 
The reconstructed image appears
sufficiently clean for direct reconstruction, 
so I will first attempt STEM2CIF 
without template retrieval.

[TOOL_CALL] stem2cif_tool
{"img_path": ..., "elements": [...]}

[TOOL_RESPONSE] stem2cif_tool
{success: true, cif_path: ...}

[TOOL_CALL] validation_tool
{"cif_path": ...}

[TOOL_RESPONSE] validation_tool
{success: true, 
converged: true, stress_xx_GPa: ...}

[ANSWER]
The direct path succeeded.
Template retrieval was not 
invoked in this case.
\end{verbatim}
}

\paragraph{Overall sensitivity to the choice of LLM/VLM agent.}
We replace the agent backbone with Qwen3, GPT-4o, and Qwen-VL (72B) and evaluate on a stratified random sample of 230 cases.
Performance remains stable, indicating that gains stem from the framework rather than a particular backbone.

% 简单表格单列即可
\begin{table}[h]
\centering
\caption{Formation-energy MAE (meV/atom) across tiers for different agents.}
\label{tab:mae_llm_vlm}
\footnotesize
\begin{tabular}{lcccc}
\toprule
\textbf{Method} & \textbf{Tier 1} & \textbf{Tier 2} & \textbf{Tier 3} & \textbf{Avg.} \\
\midrule
Qwen-VL (72B)   & 261.6 & 308.5 & 446.1 & 313.8 \\
Qwen3 agent     & 245.2 & 306.2 & 435.1 & 310.5 \\
GPT-4o agent    & 240.7 & 304.2 & 435.0 & 308.5 \\
DeepSeekV3      & 243.8 & 305.3 & 433.5 & 309.6 \\
\bottomrule
\end{tabular}
\end{table}

\begin{table}[h]
\centering
\caption{Structural accuracy metrics across tiers.}
\label{tab:struct_acc_llm_vlm}
\footnotesize
\setlength{\tabcolsep}{3pt} % 减少列间距
\begin{tabular}{clccc}
\toprule
\textbf{Tier} & \textbf{Method} & \textbf{RMSD (\AA)} & \textbf{C.C. (\%)} & \textbf{S.S. (\%)} \\
\midrule
1 & Qwen-VL      & $0.12 \pm 0.02$ & 100 & 100 \\
  & DeepSeekV3   & $0.11 \pm 0.02$ & 100 & 100 \\
\midrule
2 & Qwen-VL      & $0.11 \pm 0.03$ & 83.1 & 79.2 \\
  & DeepSeekV3   & $0.11 \pm 0.02$ & 85.9 & 79.8 \\
\midrule
3 & Qwen-VL      & $0.12 \pm 0.02$ & 58.3 & 58.3 \\
  & DeepSeekV3   & $0.11 \pm 0.03$ & 66.7 & 66.7 \\
\bottomrule
\end{tabular}
\end{table}

% 【修改点4】宽表格必须跨栏 (table*)
\begin{table*}[t]
\centering
\caption{Tool invocation, success, and retry statistics of different agents in the AutoMat toolchain.}
\label{tab:tool_stats}
\small
\begin{tabular}{l l r r r r r}
\toprule
\textbf{LLM Agent} & \textbf{Tool} & \textbf{Total Calls} & \textbf{Callbacks} & \textbf{Successes} & \textbf{Retries} & \textbf{Success Rate (\%)} \\
\midrule
Qwen-VL (72B) & Denoising           & 293,417 & 259,217 & 259,217 &  34,200 & 88.34 \\
              & Property Prediction & 165,594 & 121,153 &  68,532 &  97,062 & 41.39 \\
              & STEM2CIF            & 314,001 & 252,801 & 133,282 & 180,719 & 42.45 \\
              & Template Matching   & 258,987 & 205,762 & 205,762 &  53,225 & 79.45 \\
\midrule
GPT-4o        & Denoising           &  58,053 &  56,950 &  56,950 &   1,103 & 98.10 \\
              & Property Prediction &  38,705 &  36,563 &  36,563 &   2,142 & 94.46 \\
              & STEM2CIF            &  46,600 &  43,507 &  43,507 &   3,093 & 93.36 \\
              & Template Matching   &  59,506 &  55,951 &  55,951 &   3,555 & 94.02 \\
\bottomrule
\end{tabular}
\end{table*}

\paragraph{Decision-making flexibility.}
AutoMat is an agentic controller rather than a fixed pipeline: it dynamically decomposes subtasks and adapts tool invocation.
If a tool fails, rollback-and-retry is triggered until success or termination.

% ------------------------------ Failure Analysis ------------------------------
\subsection{Failure Analysis}
\label{appendix:failure_analysis}

We categorize failure cases into:
\begin{itemize}[leftmargin=*, itemsep=2pt, topsep=2pt]
  \item \textbf{Template-retrieval failures (39.3\%).}
  The correct structure is absent from the template database. \emph{Representative case:} Tier~1, \texttt{2dm-5936} (Se-oxide vs.\ nitride). High-$Z$ Se columns resemble N-like contrast.

  \item \textbf{Downstream failures (60.7\%).}
  The correct template is retrieved, but downstream reconstruction fails due to projection effects or extreme contrast imbalance.
  See Table~\ref{tab:failure_modes}.
\end{itemize}

% 【修改点5】使用 table* 让 Failure Modes 表格跨栏，否则 tabularx 在窄栏会非常难看
\begin{table*}[t]
\centering
\caption{Breakdown of downstream failure sub-modes despite correct template retrieval.}
\label{tab:failure_modes}
\small
\begin{tabularx}{0.9\linewidth}{l c X X}
\toprule
\textbf{Sub-mode} & \textbf{Share} & \textbf{Core Problem} & \textbf{Representative Case} \\
\midrule
Projection overlap & 40\% &
Atoms in different $z$-layers project to the same $(x,y)$, confusing structural relaxation &
Tier~2, \texttt{2dm-1014}: B/I overlap; I dominates iDPC, B is lost \\
Extreme $Z$-contrast & 20.7\% &
Heavy elements dominate contrast and mask lighter atoms, treated as noise &
Tier~3, \texttt{2dm-5199}: U contrast hides O/F, leading to $>3$~eV energy error \\
\bottomrule
\end{tabularx}
\end{table*}

% ---------------------- Real STEM (ZSM-5) Reconstruction ----------------------

% Put the wide figure FIRST so it can sit at the TOP of the page (two-column mode).
\begin{figure*}[!t]
  \centering
  % Control height to avoid float-only page centering / awkward whitespace.
  \includegraphics[width=\textwidth,height=0.32\textheight,keepaspectratio]{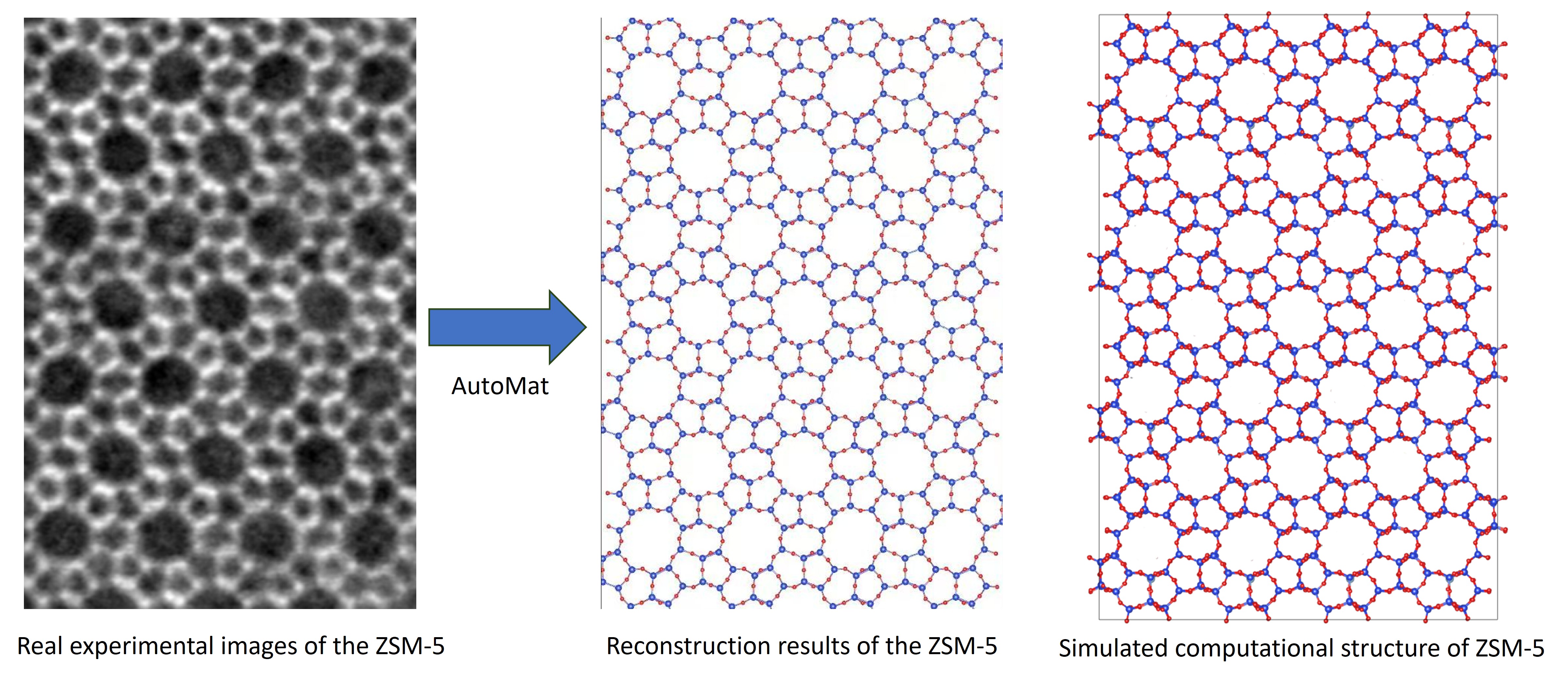}
  \caption{\textbf{AutoMat reconstruction on real iDPC-STEM images of ZSM-5 zeolite.}
  (Left) Experimental iDPC-STEM image showing characteristic ten-membered-ring pores.
  (Middle) Structure reconstructed by AutoMat from the real image.
  (Right) DFT-relaxed reference model of ZSM-5.}
  \label{fig:zsm5_real_recon}
\end{figure*}

\subsection{Real STEM Image Reconstruction on ZSM-5}
\label{appendix:real_stem_zsm5}

Due to experimental limitations, our current setup does not yet support systematic preparation and iDPC-STEM imaging of 2D materials. To provide an informative evaluation on real experimental data, we choose ZSM-5 zeolite as a representative test case.

\noindent\textbf{ZSM-5 iDPC-STEM imaging experiments.}
Experiments were carried out on a Cs-corrected STEM (FEI Titan Cubed Themis G2 300) equipped with a four-quadrant detector at 300~kV. Sheet-like ZSM-5 single crystals were ultrasonically dispersed in pyridine or thiophene for 1--2~h, deposited onto microgrids, and heated at 130~$^\circ$C.

\noindent\textbf{AutoMat reconstruction on real ZSM-5 images.}
AutoMat recovers the framework topology and continuous ten-membered-ring channels consistent with the reference. After alignment, framework atomic deviations are within $\sim 0.2$~\AA{}, indicating reasonable structural fidelity under realistic experimental noise.

% ----------------------- Limitations and Future Directions -----------------------
\subsection{Limitations and Future Directions}
\label{appendix:limitations}

AutoMat currently targets \textbf{2D monolayer systems imaged with iDPC-STEM}, which provides a stable and interpretable testbed but also introduces several important limitations.

\begin{itemize}[leftmargin=*, itemsep=2pt, topsep=2pt]
  \item \textbf{Limited real-data validation.}
  Although we include a real iDPC-STEM case on ZSM-5 in Appendix~\ref{appendix:real_stem_zsm5}, this experiment should be interpreted as an \emph{initial feasibility demonstration} rather than a substitute for systematic validation on multiple real 2D monolayer materials. High-quality raw iDPC-STEM data for 2D materials are currently scarce in public repositories, and acquiring new large-field, dose-limited experimental data requires substantial sample-preparation effort, aberration-corrected STEM beamtime, and repeated calibration. Broader real-material validation therefore remains an important next step.

  \item \textbf{2D scope and single-projection depth ambiguity.}
  The current pipeline assumes a single iDPC-STEM projection. Since iDPC-STEM mainly reflects the projected electrostatic potential integrated along the beam direction, a single image does not preserve sufficient depth information to uniquely recover a general 3D atomic structure. This makes reconstruction of bulk 3D crystals intrinsically ill-posed from a single projection. By contrast, 2D monolayer materials have limited extent along the $z$ axis, which substantially reduces projection ambiguity and makes them a physically well-posed starting point.

  \item \textbf{Element classification uncertainty.}
  Element assignment relies on contrast cues and template priors; similar-$Z$ elements and low-dose conditions increase confusion.

  \item \textbf{Template retrieval sensitivity.}
  Mismatches in pixel scale, defocus, and aberrations reduce robustness.

  \item \textbf{Heuristic agent policy.}
  The current policy uses state-dependent rules; it may not remain optimal as the tool pool grows.
\end{itemize}

\paragraph{Future directions.}
We plan to address these limitations along four main directions.
First, for \textbf{real-data robustness}, we will expand real-image evaluation to multiple monolayer materials with known CIFs, enlarge the template library to cover a broader range of defocus and aberration conditions, and introduce simple calibration steps for pixel scale and imaging-condition drift.
Second, for \textbf{3D extension}, we are exploring two concrete routes: \emph{electron tomography}, where multiple tilt-angle projections are combined for 3D reconstruction, and \emph{4D-STEM}, where diffraction information at each probe position may provide richer depth-sensitive structural signals.
Third, we plan to adopt \textbf{contrast-sensitive recognition} and fuse complementary modalities such as \textbf{EELS/EDX} to improve species discrimination under overlap or similar-$Z$ conditions.
Finally, we will explore \textbf{learnable policy optimization} and more robust retrieval strategies, including \textbf{pre-retrieval calibration}, to improve controller scalability and robustness under broader experimental conditions.

\end{document}